\documentclass{article}

\usepackage[preprint]{neurips_2019}

\usepackage[utf8]{inputenc} 
\usepackage[T1]{fontenc}    
\usepackage{hyperref}       
\usepackage{url}            
\usepackage{booktabs}       
\usepackage{amsfonts}       
\usepackage{nicefrac}       
\usepackage{microtype}      
\usepackage{graphicx}
\usepackage{amsmath}  
\usepackage{amssymb}   
\usepackage{amsthm}
\usepackage{wrapfig}  
\usepackage{algorithm}
\usepackage{algpseudocode}
\usepackage{xcolor}
\newtheorem{theorem}{Theorem}[section]

\newtheorem{lemma}[theorem]{Lemma}
\newtheorem{definition}{Definition}[section]
\newtheorem{example}{Example}[section]
\usepackage{xspace}
\usepackage{subfig}
\usepackage[english]{babel}
\usepackage{blindtext}
\newcommand{\mapname}{adversarial map\xspace}

\graphicspath{{images/}}

\title{Identifying Classes Susceptible to Adversarial Attacks}
\author{%
	Rangeet Pan\\
	Department of Computer Science\\
	Iowa State University\\
	\texttt{rangeet@iastate.edu} \\
	\And
	Md Johirul Islam\\
	Department of Computer Science\\
	Iowa State University\\
	\texttt{mislam@iastate.edu} \\
	\And
	Shibbir Ahmed\\
	Department of Computer Science\\
	Iowa State University\\
	\texttt{shibbir@iastate.edu} \\
	\And
	Hridesh Rajan\\
	Department of Computer Science\\
	Iowa State University\\
	\texttt{hridesh@iastate.edu} \\
}

\begin{document}

\maketitle

\begin{abstract}
	Despite numerous attempts to defend deep learning based image classifiers, 
	they remain susceptible to the adversarial attacks. 
	This paper proposes a technique to identify {\em susceptible classes}, those
	classes that are more easily subverted.
	To identify the susceptible classes we use distance-based measures and apply
	them on a trained model.
	Based on the distance among original classes, we create mapping among original
	classes and adversarial classes that helps to reduce the randomness of a model
	to a significant amount in an adversarial setting.
	We analyze the high dimensional geometry among the feature classes and identify
	the $k$ most susceptible target classes in an adversarial attack. 
	We conduct experiments using MNIST, Fashion MNIST, CIFAR-10 (ImageNet and
	ResNet-32) datasets. Finally, we evaluate our techniques in order to determine
	which distance based measure works best and how the randomness of a model
	changes with perturbation. 	
\end{abstract}
\section{Introduction}
\label{sec:intro}
Protecting against adversarial attacks has become an important concern for 
machine learning (ML) models since an adversary can cause a model to misclassify
an input
with high confidence by adding small perturbation~\citep{goodfellow2014explaining}. 
A number of prior works ~\citep{30papernot2016limitations, franceschi2018robustness, 2yan2018deep,  6pang2018towards, 9zheng2018robust, 8tao2018attacks, 20tian2018detecting, 21goswami2018unravelling,3fawzi2018adversarial,12peck2017lower,28papernot2016distillation} have tried to understand the characteristics of adversarial attacks. 
This work focuses on adversarial attacks on 
deep neural networks (DNN) based image classifiers.


\textbf{Our Contributions.}
Our work is driven by the two fundamental questions.
Can an adversary fool all classes equally well? 
If not, which classes are {\em susceptible} to adversarial attacks more so than others?
Identifying such classes can be important for developing better defense mechanisms.
We introduce a technique for identifying the top $k$ susceptible classes. 
Our technique analyzes the DNN model to understand the high dimensional 
geometry in the feature space.
We have used four different distance-based measures (\emph{t-SNE, N-D Euclidean, 
	N-D Euclidean Cosine}, and \emph{Nearest Neighbor Hopping} distance) for 
understanding the feature space.
To determine the top $k$ susceptible classes, we create an \mapname, 
which requires the distance in feature space among classes as input and 
outputs a mapping of probable adversarial classes for each actual class. 
To create \mapname, we introduce the concept of the forbidden distance 
i.e., the distance measured in high dimension which describes the capability 
of a model to defend an adversarial attack. 

%

We conduct experiments on FGSM attack using MNIST \citep{lecun1998gradient}, 
Fashion MNIST \citep{xiao2017fashion}, and CIFAR-10 \citep{krizhevsky2009learning} 
(ImageNet~\citep{krizhevsky2012imagenet} and ResNet-32 \citep{he2016deep}) datasets 
to evaluate our technique. 
Finally, we compare our results with cross-entropy (CE)\citep{goodfellow2016deep} 
and reverse cross-entropy (RCE)\citep{3fawzi2018adversarial} based training 
techniques that defend against adversarial attack. 
Our evaluation suggests that in comparison to the previous state-of-the-art training 
based techniques, our proposed approach performs better and 
does not require additional computational resources.

Next, we describe related works. $\S\ref{sec:approach}$ describes our methodology, $\S\ref{sec:eval}$ describes detailed results with the experimental setup, and $\S\ref{sec:conclusion}$ concludes.

\section{Related Work}
\label{sec:related}
The work on adversarial frameworks~\citep{15goodfellow2014generative} can be 
categorized into attack and defense related studies. 

{\bf Attack-related studies.\ } Several studies have crafted attacks on ML models, e.g. 
FGSM~\citep{goodfellow2014explaining}, CW~\citep{carlini2017towards}, JSMA~\citep{30papernot2016limitations}, 
Graph-based attack~\citep{17zugner2018adversarial}, 
attack on stochastic bandit algorithms~\citep{jun2018adversarial}, 
black-box attacks~\citep{liu2016delving,papernot2017practical}, etc. \citep{5elsayed2018adversarial} studies transferability of attacks, and \citep{23athalye2017synthesizing} surveys the kinds of attack using synthesizing robust adversarial examples for any classifier. 

{\bf Defense techniques (our work fits here).\ } 
These works are primarily focused on 
improving robustness~\citep{2yan2018deep, 6pang2018towards, 9zheng2018robust}, 
detecting adversarial samples ~\citep{8tao2018attacks, 9zheng2018robust}, 
image manipulation ~\citep{20tian2018detecting, 21goswami2018unravelling}, 
attack bounds ~\citep{3fawzi2018adversarial,12peck2017lower}, 
distillation \citep{28papernot2016distillation}, 
geometric understanding \citep{30papernot2016limitations}, etc. 
There are other studies on the geometrical understanding of 
adversarial attack ~\citep{30papernot2016limitations, franceschi2018robustness, schmidt2018adversarially}.
Papernot {\em et al.}'s~\citep{30papernot2016limitations} work is closely related 
to ours, where the authors built a capability-based adversarial saliency map 
between benign class and adversarial class to craft perturbation in the input. 
In contrast, we utilize distance-based measures to understand a DNN model 
and detect $k$-susceptible target classes. 
\citep{franceschi2018robustness} utilizes the decision boundary to understand 
the model and the authors have observed a relationship between the decision 
boundary and the Gaussian noise added to the input. 
\citep{18Gilmer} has conducted a similar study to understand how decision 
boundary learned by a model helps to understand high dimensional data and 
proposed the bound over the error of a model. 
Our approach finds the relation of high dimensional geometry with adversarial 
attacks and identifies $k$ susceptible classes. 

\section{Our Approach: Identifying Susceptible Classes}
\label{sec:approach}
We use distances ($\S\ref{subsec:distcal}$)
to create \mapname ($\S\ref{subsec:attackmap}$) and use it to 
pick susceptible classes ($\S\ref{subsec:lowcostdet}$).

\subsection{Terminology}
\label{subsec:terms}
This study concentrates on the feed-forward DNN classifiers. 
A DNN can be represented as a function $f(X,p): \mathbb{R}^N \rightarrow \mathbb{R}^n$,
where $p$ is the set of tunable parameters, $X$ is the input, $n$ is the number of labeled
classes and $N$ is the number of features and $X\in \mathbb{R}^N$. 
In this study, the feature space for a model has been represented using $\mathbb{R}^N$.
The focus of the paper is to understand the high dimensional geometry to identify
susceptible target classes for a model. 
We calculate distance $d(c_{i},c_{j})$ between two classes $c_i$ and $c_j$, where $i\ne j$.
We  utilize four different distance-based measures and compare them. 
In adversarial setting, $f(X,p)\rightarrow f(X+\eta,p)$, where $\eta$ is the perturbation
added to the input which causes, $f(X,p)\ne f(X+\eta,p)$. 
Our assumption in this study is that $f(X+\eta,p)$ depends on the $d(c_i,c_j)$ of a model.
To compare different distances, we calculate the randomness in a model using the entropy.
We define the entropy of a model $M$ as $e_M$ where,
\begin{equation}
e_M=-\Sigma P(x_i)\log(P(x_i))
\end{equation}
$P(x_i)$ denotes the probability of input $x_i$, which has been misclassified to class $c_j$
given the actual label $c_i$. 
In this study, we use terms e.g., actual class and adversarial class, which represents the
label of a data point predicted by a model and the label after a model has gone through 
an attack respectively. 
We introduce a term \emph{forbidden distance} as $D$, a measured distance which
provides the upper bound of displacement of data points in $\mathbb{R}^N$. 
In this context, displacement represents the distance between the adversarial class 
and the actual class.
In this study, we have conducted our experiment using the \emph{Fast Gradient Sign
	Method (FGSM)}~\citep{goodfellow2014explaining}.  
Here, we have chosen single attack based on the 
\emph{Adversarial transferable property}~\citep{30papernot2016limitations, papernot2016towards}, 
which defines that adversarial examples created for one model are highly 
probable to be misclassified by a different model.
\subsection{Hypothesis ($H_0$)}
\label{subsec:hypothesisl}
According to \textit{linearity hypothesis} proposed in~\citep{goodfellow2014explaining},
there is still a significant amount of linearity present in a model even though a DNN model utilizes non-linear transformation. The primary reason behind this is the usage of LSTM~\citep{hochreiter1997long}, ReLu~\citep{jarrett2009best,glorot2011deep}, etc, which possess a significant amount of linear components to optimize the complexity. 
Here, we assume that the input examples can be misclassified to neighboring classes 
in the $\mathbb{R}^N$ during adversarial attack.
\subsection{Distance Calculation}
\label{subsec:distcal}
\paragraph{\textbf{Calculation of t-SNE distance}}
To understand $\mathbb{R}^N$ representation, we utilize t-SNE \citep{maaten2008visualizing} dimension reduction technique. t-SNE uses the Euclidean distance of data points in $N$ dimension as input and converts it into the probability of similarity, $p_{j|i}=\frac{\exp (-||x_{i}-x{j}||^{2}/2\sigma_{i}^{2})}{\Sigma_{i\ne j}\exp (-||x_{i}-x{j}||^{2}/2\sigma_{i}^{2})}$
where, $p_{j|i}$ represents the probability of similarity between two input data points $i$ and $j$ in $\mathbb{R}^N$. We calculate the distance $d(c_i,c_j)$ based on the $p_{j|i}$. We convert $N$-dimensional problem to a $2$-dimensional problem. In this process, we do not consider the error due to the curse of dimensionality \citep{indyk1998approximate}. In 2-D feature space, we have the co-ordinate $x_{i}(\vec{x},\vec{y})$ for a data point $x_i$ and we calculate the center of mass $cm_i$, where $i$ is the class. Here, mass of each point $x_i$, is assumed to be unit, then center of mass $cm_i$ represents,
$cm_i=\frac{c_ix_{1}+c_ix_{2}+...+c_ix_{n}}{|c_i|}$, where,
$c_ix_j$ represents the $x_j$ data point of class $c_i$. 
\paragraph{\textbf{Calculation of N-D Euclidean Distance}}
Furthermore, we calculate the $N$-dimensional Euclidean distance between two data points. Each data point can be represented by a feature vector $\vec{F}=\{c_i\vec{x_j}(1), c_i\vec{x_j}(2),..., c_i\vec{x_j}(N)\}$, where $c_i\vec{x_j}(k)$ is the $k^{th}$ vector component of data point $x_j$ of class $c_i$. Here, $c_i\vec{x_j}(k)$ has been represented as a coordinate in $\mathbb{R}^N$. We calculate the center of mass similar to the t-SNE based approach. The main difference is the calculated center of mass $cm_i$ is a vector of $N$ coordinates. Thus, $\forall i\ne j$, distance $d(c_i,c_j)$ can be calculated as, 
\begin{equation}
d(c_i,c_j) = \sqrt{(cm_j(1)-cm_i(1))^{2}+...+(cm_j(N)-cm_i(N))^{2})}
\end{equation}
\paragraph{\textbf{Calculation of N-D Euclidean Cosine Distance}}
We use the $N$ dimensional angular distance as our next measure. In this process, we calculate the $N$-dimensional Euclidean distance similar to the prior technique. In $N$-dimension, the angular similarity among the center of mass of classes $cm_i$, $cm_j$ can be calculated as,
$\theta_{cm_i,cm_j}=\frac{\vec{cm_i}.\vec{cm_j}}{||\vec{cm_i}|.||\vec{cm_j}||}$, 
where $||\vec{cm_j}||$ is the magnitude of the $cm_j$ vector. We leverage the angular similarity and calculate the $N$-dimensional Euclidean angular distance between center of mass of two classes using the following equation,
\begin{equation}
d(c_i,c_j) = \sqrt{(cm_j(1)-cm_i(1))^{2}+...+((cm_j(n)-cm_i(n))^{2})}*\cos\left(\frac{\vec{cm_i}.\vec{cm_j}}{||\vec{cm_i}|.||\vec{cm_j}||}\right)
\end{equation}
\vspace{-25pt}
\begin{algorithm}[H]
	\caption{Hopping Distance}
	\label{algo:hopdist}
	\begin{algorithmic}[1]
		\Procedure{hopDistance ($X_i,X_j$)}{}
		\State $V_{child}=\{$ Neighbor of $X_i\}$ \label{algo2:l1}
		\State $distance=0$; \label{algo2:l2}
		\State Mark $X_i$ visited;\label{algo2:l3} $V_{list}=NULL$\label{algo2:l41}
		\If {($X_i\ne X_j$)}\label{algo2:l5}
		\While {$X_j\notin V_{list}$}\label{algo2:l6}
		\State distance=distance+1\label{algo2:l7}
		\State $V_{expand}=NULL$\label{algo2:l8}
		\For {each $v\in V_{child}$}\label{algo2:l9}
		\If{($v$ not visited)}\label{algo2:l10}
		\State $v_x=\{$Neighbor of $v\}$\label{algo2:l11}
		\State $V_{expand}=V_{expand}\cup v_x$\label{algo2:l12}
		\State Mark $v$ visited
		\EndIf\label{algo2:l13}
		\EndFor\label{algo2:l14}
		\State $V_{list}=V_{list}\cup V_{expand}$\label{algo2:l15}
		\State $V_{child}=V_{expand}$\label{algo2:l16}
		\EndWhile\label{algo2:l17}
		\EndIf\label{algo2:l18}
		\State \Return distance\label{algo2:l19}
		\EndProcedure\label{algo2:l20}
	\end{algorithmic}	
\end{algorithm}
\paragraph{\textbf{Calculation of Nearest Neighbor Hopping Distance}}
Here, we use the nearest neighbor algorithm to understand the behavior of an adversarial attack. For distance calculation, we develop an algorithm which computes the hopping distance \ref{algo:hopdist} between two classes. Initially, we calculate $k$ nearest neighbors for each data point. Due to \emph{Reflexive} property, a data point is inevitably neighbor to itself, which approves $(k-1)$ different neighbors for a data point. The boundary learned by the nearest neighbor algorithm distinguishes classes by dividing into clusters. Then, nearest neighbors will belong to the same class for most of the data points except the data points located near the boundary. We leverage that information and compute the classes nearest to a particular class.
\begin{example}
	\label{ex:1}
	Let us assume the points in class $c_1 = \{x_1, x_2, x_3, x_4\}$, $c_2 = \{y_1, y_2, y_3, y_4\}$, $c_3 = \{z_1, z_2, z_3, z_4\}$. 
	For example, we find the closest point of $x_1$ outside $c_1$ is $z_1$, of $x_2$ is $y_1$, of $x_3$ is $y_2$, $x_4$ is $y_3$. 
	As depicted in Figure \ref{fig:attackmapexample}(b), we observe that $\frac{3}{4}$ of the data points in $c_1$ have their closest neighbors in $c_2$. So, we can say that $c_2$ shares more boundary with $c_1$ and is the closest neighboring class to $c_1$. 
\end{example}

\textbf{Hopping Distance} computes how many hops a data point needs to travel to reach the closest data point in the target class. From the nearest neighbor algorithm, we get the $(k-1)$ unique neighbors to each data point. Algorithm \ref{algo:hopdist} takes the output from the $k$ nearest neighbor, the actual predicted data point and the misclassified label as input. This problem has been converted to a problem of tree generation from lines \ref{algo2:l1} - \ref{algo2:l3} . From lines \ref{algo2:l5} - \ref{algo2:l20}, we expand the tree when a new neighbor has been found and traverse using BFS. Finally, we calculate the depth of the expanded tree to calculate the minimum distance that a data point has to travel to reach the misclassified class in $\mathbb{R}^N$. This algorithm utilizes the same time and space complexity as BFS does, which is $O(V)$ for time and $O(V)$ for space, as in the worst case, we need to traverse all the neighbors ($|V|$) for an actual class.  
We also calculate the forbidden distance based on the average hopping distance ($D$) for a model. In the Eq.\ref{eq:Hop}, $n$ and $|X|$ denote the total number of classes and data points respectively.
We use Eq.\ref{eq:Hop} for both calculating the forbidden distance ($F_d$) and also the average displacement of data points in $\mathbb{R}^N$ under an attack. For calculating the later, $x_i$ is the actual class and $x_j$ is the adversarial class. In order to create the \mapname, we compute a matrix storing the distance among all classes ($S_D$) using Eq. \ref{eq:avgHop}, where $x_k$ is the total number of data points in class $c_k$.
\begin{equation}
\label{eq:Hop}
D=\frac{\Sigma_{\forall (x_i \in c_k,x_j \in c_l,k,n \in \{1,2,...,n\}, i\ne j)}HopDistance(x_i,x_j)}{|X|}
\end{equation}
\begin{equation}
\label{eq:avgHop}
S_D=\bigcup_{l,k\in{1,2,...,n}}\frac{\Sigma_{\forall (x_i \in c_k,x_j \in c_l,k,n \in \{1,2,...,n\}, i\ne j)}HopDistance(x_i,x_j)}{|x_k|}
\end{equation}
\begin{lemma}
	$\forall x \in c_i, \forall y \in c_j,\forall z \in c_k$ if average hopping distance $d(c_i, c_j) < d(c_i, c_k)$, then $c_j$ is closer to $c_i$ than $c_k$ to $c_i$ i.e., the distance of center of mass $d(cm_i, cm_j) < d(cm_i, cm_k)$.
	\begin{proof}
		$\frac{\sum_{l,m} {d(x_l, y_m)}}{|c_i * c_j|} < \frac{\sum_{l,n} {d(x_l, z_n)}}{|c_i * c_k|}$. Without the loss of generality, we can say that, $
		\sum_{l,m} {d(x_l, y_m)} < \sum_{l,n} {d(x_l, z_n)}$. 
		As center of mass will always be within the polygon surrounding a class, without loss of generality we assume all the $y_m$ are at the same location $p$ and all the $z_n$ are at the same location $q$.
		\begin{align*}
		\sum_{l,m} {d(x_l, y_m)} &< \sum_{l,n} {d(x_l, z_n)} \\
		\implies
		\frac{|x_1 - p| + |x_2 - p| + ... + |x_l - p|}{|c_i*c_j|} &< \frac{|x_1 - q| + |x_2 - q| + ... + |x_l - q|}{|c_i*c_k|}\\
		\implies
		\frac{cm_i}{|c_j|} - \frac{cm_j}{|c_i|} &< \frac{cm_i}{|c_k|} - \frac{cm_k}{|c_i|}
		\end{align*}
		Assuming the balanced dataset, $|c_i| \sim |c_j| \sim c_k$,
		$\implies cm_i - cm_j < cm_i - cm_k$
		So, the center of mass of  $c_j$ is closer to $c_i$ than the center of mass of $c_k$ to $c_i$.
	\end{proof}
\end{lemma}
\begin{lemma}
	In $\mathbb{R}^N$, if a class $c_i$ has been misclassified to a closer class $c_j$, the entropy $e_M$ will decrease.
	\begin{proof}
		The entropy $e_M=-\Sigma P(x_i)\log(P(x_i))$. With the increase of $P(x_i)$, $\log(P(x_i))$ also increases. So, we can say that $e_M \propto -P(x_i) \implies e_M \propto \frac{1}{P(x_i)}$ . From $H_0$ (\S\ref{subsec:hypothesisl}), we assume that if a class $c_i$ is close to class $c_j$, we allocate a higher probability to $P(x_i)$. So, if classes are mostly misclassified to the closer one, the entropy of the entire model will decrease.
	\end{proof}
\end{lemma} 
\subsection{Adversarial Map}
\label{subsec:attackmap}
%
%
%
\begin{figure}
	\vspace{-10pt}
	\centering
	\subfloat[]{\includegraphics[trim=2 1 5 2,clip,width=0.4\linewidth]{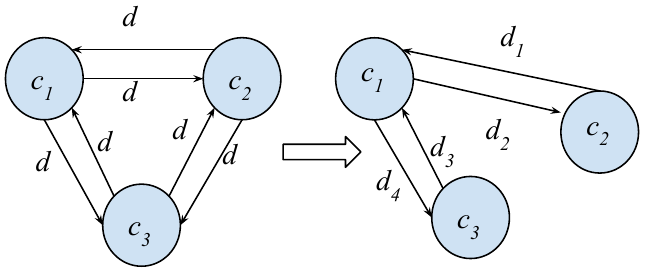}}
	\hspace{20pt}
	\subfloat[]{\includegraphics[trim=5 2 5 0,clip,width=0.35\linewidth]{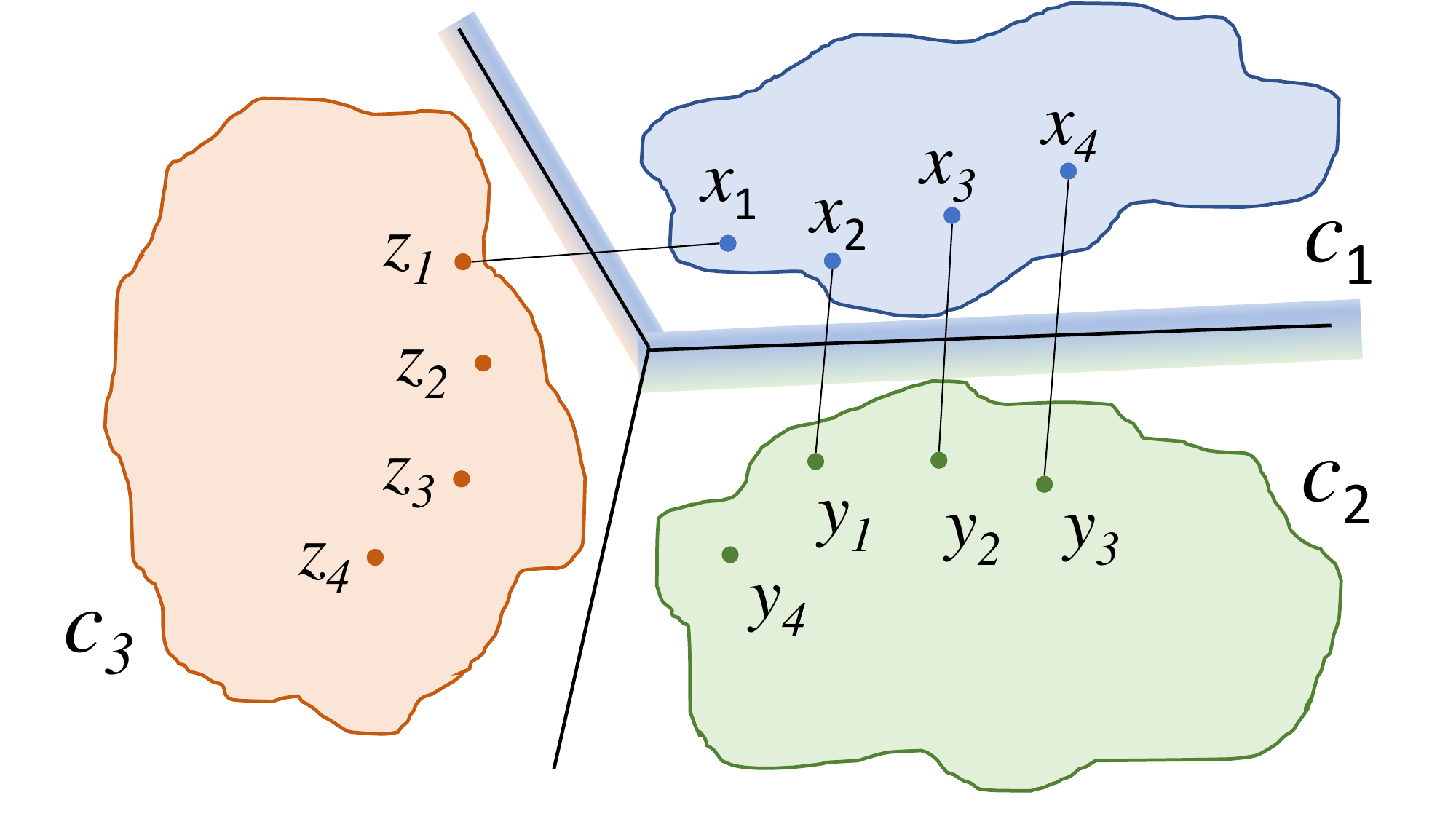}}
	\caption{(a) Creation of Adversarial Map. (b) Class $c_1$ shares more boundary with $c_2$ than $c_3$ as $\frac{3}{4}$ of the data points in $c_1$ have their closest neighbors in $c_2$.}
	\label{fig:attackmapexample}
\end{figure}

\begin{wrapfigure}{L}{0.5\linewidth}
	\vspace{-15pt}
	\begin{minipage}{0.5\textwidth}
		\begin{algorithm}[H]
			\caption{Create Adversarial Map}
			\label{algo:createmap}
			\begin{algorithmic}[1]
				\Procedure{createMap ($S_D$)}{}
				\State $G \gets $ empty graph \label{algo1:l1}
				\For{each $U, V \in$ $S_D$}\label{algo1:l2}
				\If{($d(U,V) \le F_d$)} \label{algo1:l3}
				\State $G \gets G \cup edge(U,V,d(U,V))$ \label{algo1:l4}
				\EndIf \label{algo1:l5}
				\EndFor \label{algo1:l6}
				
				\Return G \label{algo1:l7}
				\EndProcedure
			\end{algorithmic}	
		\end{algorithm}	
	\end{minipage}
\end{wrapfigure}
\begin{definition}
	\textbf{Forbidden distance ($F_d$): } When a model encounters an adversarial attack, each input class requires to travel a certain distance in $\mathbb{R}^N$ to accomplish the attack. Based on the attack type, maximum distance changes. We call this forbidden distance ($F_d$) as beyond this distance adversarial attack will not be successful.   
	For example, to accomplish the adversarial attack given a forbidden distance of a model $F_d$ and to misclassify $c_1$ as $c_2$, distance constraint between $c_1$ and $c_2$ is $d(c_1, c_2) \le F_d$. 
\end{definition}
In this section, we describe how we create the \mapname annotated with distance to neighbors. Here, we utilize the forbidden distance while creating the \mapname.
Hypothetically, any class $c_1$ as shown in Figure \ref{fig:attackmapexample}(a) can be misclassified to any other class by traveling the same distance $d$.
But our hypothesis is that every attack has a limitation. A data point in $c_i$ might need to travel different distance for misclassifying to different classes $c_k$, where $k \in \{1,2, ..., n\}, i \ne k$. If we represent the distance between $c_i$ and $c_k$ as $d(c_i, c_k)$ then the attack can be accomplished more easily where $d(c_i,c_k)$ is minimum.
$argmin(d(c_s, c_t)) \implies MP(c_s \rightarrow c_t)$
In the above equation if $d(c_s, c_t)$ is minimum then the attack can missclassify $c_s$ as $c_t$ represented by the function $MP$. We create this \mapname by using the distance between classes $c_i$ and $c_j$ as described in \S \ref{subsec:distcal}. Then we introduce the notion of forbidden distance $F_d$. We claim that the attack on a certain class $c_s$ can misclassify as class $c_t$ if and only if $d(c_s,c_t) \le F_d$ as mentioned in the following equation:
\begin{equation}
c_s \rightarrow c_t \iff d(c_s,c_t) \le F_d
\end{equation}
Now, we create the \mapname from the distance between different classes as depicted in Algorithm \ref{algo:createmap}, which takes the distance between different classes ($S_D$) as input. Then, different edges are added to the graph $G$ mentioned in lines \ref{algo1:l2} - \ref{algo1:l6}. Finally, the \mapname is returned in line \ref{algo1:l7}. This algorithm runs in $O(|S_D|^2)$ time and $O(n)$ space complexity.
\begin{lemma}
	The attack can misclassify a class $c_i$ only to one of its neighbors $c_j$ in \mapname.
	\begin{proof}
		Let us assume an attack has a prior knowledge of a model, training example of class $c_i$ and can misclassify to $c_k$ which is not a neighbor. 
		We know an attack can only misclassify $c_i$ as $c_k$ if and only if $d(c_i, c_k) \le F_d$. 
		According to Algorithm \ref{algo:createmap} if $d(c_i, c_k) \le F_d$ then, $c_k$ is the neighbor of $c_i$. This leads to a contradiction.     
		So the attack can only misclassify $c_i$ as one of its neighboring classes. 
	\end{proof}
\end{lemma}
\subsection{Susceptible Class Identification}
\label{subsec:lowcostdet}
Here, we use the best among four distance-based measures and identify $k$ susceptible target classes for a model. In an adversarial setting, we find the target classes which are most likely being misclassified from the actual class. Our primary hypothesis ($H_0$) claims that any class will be misclassified to the nearest class under an attack. 
In order to identify $k$ susceptible classes, we use our mapping between the actual class and adversarial class mentioned in \S \ref{subsec:attackmap}. For a particular class $c_i$, we assign weighted probability to all misclassified classes $c_j$, where $j\in {1,2, ..., n}, i\ne j$ based on the distance computed using the best distance-based measure. Higher the distance between two classes, lower the probability of one class being misclassified as another. We perform a cumulative operation on individual probability of being misclassified given the actual input label for $n$. The top $k$ classes with highest probability will be identified as the susceptible classes under an adversarial attack. 
\begin{lemma}
	Cumulative of the individual probability of adversarial classes given the actual classes determines the most susceptible classes of a model.
	\begin{proof}
		For a DNN model $M$, the data sets are categorized into $n$ classes. For each class $c_i$, there is a list of at most $(n-1)$ classes which can be close to $c_i$. For each class, we determine them based on the hypothesis $H_0$. The probability of a class $c_i$ being misclassified as $c_j$ can be determined based on the \mapname. Lesser the distance between $c_j$ and $c_i$, higher the probability of $c_i$ being misclassifed as $c_j$. So,  $ P(c_j| c_i) \propto \frac{1}{d(c_i,c_j)}$. Here, $P(c_j| c_i)$ denotes the probability of $c_i$ being misclassifed as $c_j$. As, $c_1, c_2, ... , c_n$ are all independent events, the total probability of an adversarial class $c_j$ is $P(c_j)=P(c_j|c_i) + ... + P(c_n|c_j)$ and as if class $c_i$ has been misclassified as $c_i$, we do not consider that as an adversarial effect. Hence, $P(c_i|c_i)=0$. Without the loss of generality, $P(c_j)=\Sigma_{i\in \{1, 2, ... , n\}, i\ne j}P(c_j|c_i)$. So, the probability of an adversarial class is the cumulative of individual probability of that class given all the actual classes.
	\end{proof}
\end{lemma} 
\section{Evaluation}
\label{sec:eval}
\subsection{Experimental Setup}
\label{subsec:setup}
In this study, we have used MNIST~\citep{lecun1998gradient}, 
Fashion MNIST (F-MNIST)~\citep{xiao2017fashion} 
and CIFAR-10~\citep{krizhevsky2009learning} datasets.
The number of labeled classes is 10 for each dataset. 
MNIST and F-MNIST contains 60,000 training images and 10,000 test images. 
Both train and test dataset are equally partitioned into 10 classes. 
Each class has 6,000 training and 1,000 test images. 
CIFAR-10 contains 50,000 training images and 10,000 test images. 
We have worked on one model each for MNIST and F-MNIST with 
accuracy 98\% and 89\% respectively whereas, for CIFAR-10, 
we have performed our experiment on two models, Simplified \emph{ImageNet}~\citep{krizhevsky2012imagenet} and 
\emph{ResNet-32}~\citep{he2016deep} with accuracy 72\% and 82\% respectively. 
For crafting FGSM attack, we have utilized the 
Cleverhans\citep{papernot2016cleverhans} library. 
We have experimented using four distance-based measures on each 
dataset with the label of each data point predicted by the model. 
We have run our susceptible class detection on the entire dataset 
with FGSM attack for each model and determine $k$-susceptible target classes. 
For all experiment with variable perturbation, with change the $\epsilon=0.01$ for each simulation and run the similation from $\epsilon=0.01$ to $\epsilon=2.0$. Hence, 20 simulations have been executed for each experiment.
%
\begin{figure}
	\centering
	\subfloat[]{\includegraphics[trim=10 1 15 15,clip,width=0.25\linewidth, height=.23\linewidth]{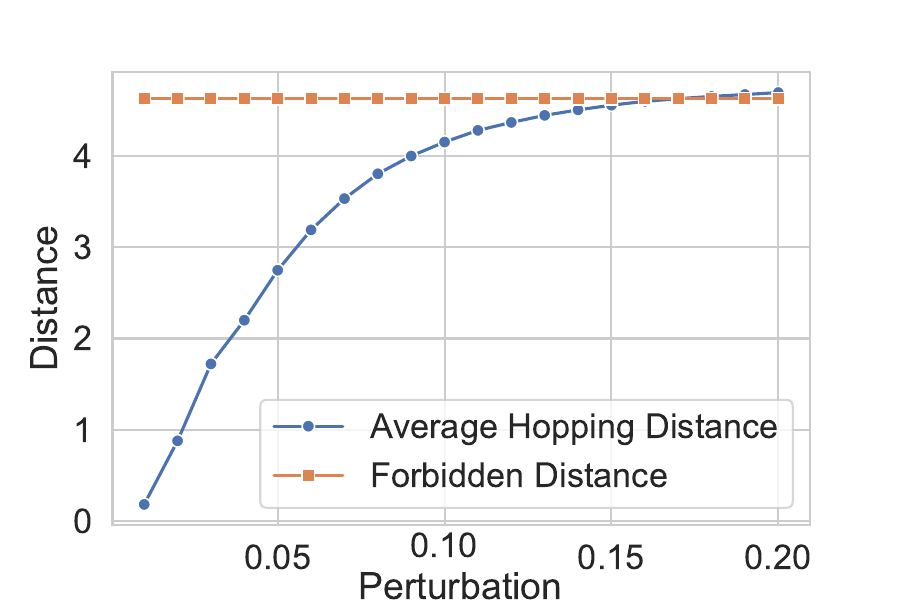}} 
	\subfloat[]{\includegraphics[trim=10 1 15 15,clip,width=0.25\linewidth, height=.23\linewidth]{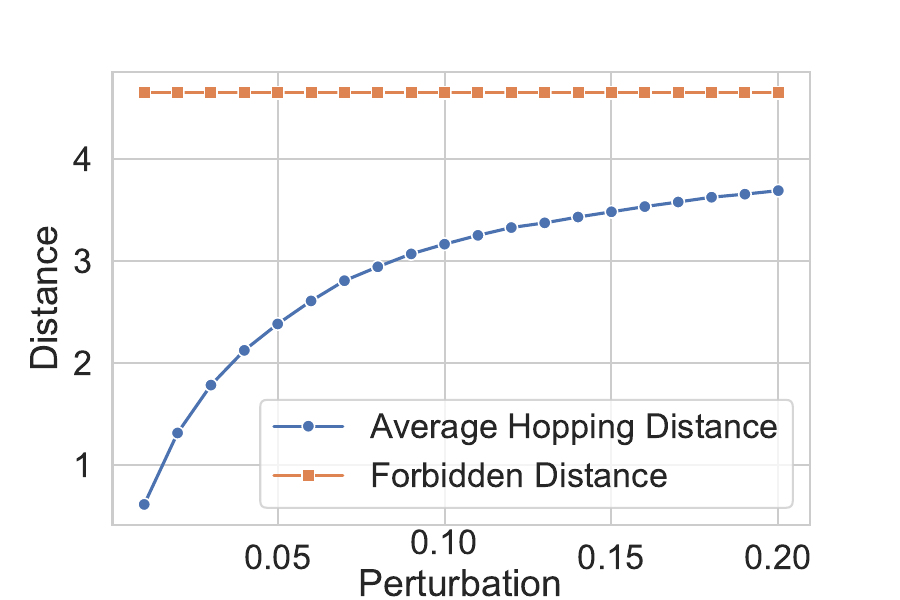}}
	\subfloat[]{\includegraphics[trim=0 1 15 15,clip,width=0.25\linewidth,height=.23\linewidth]{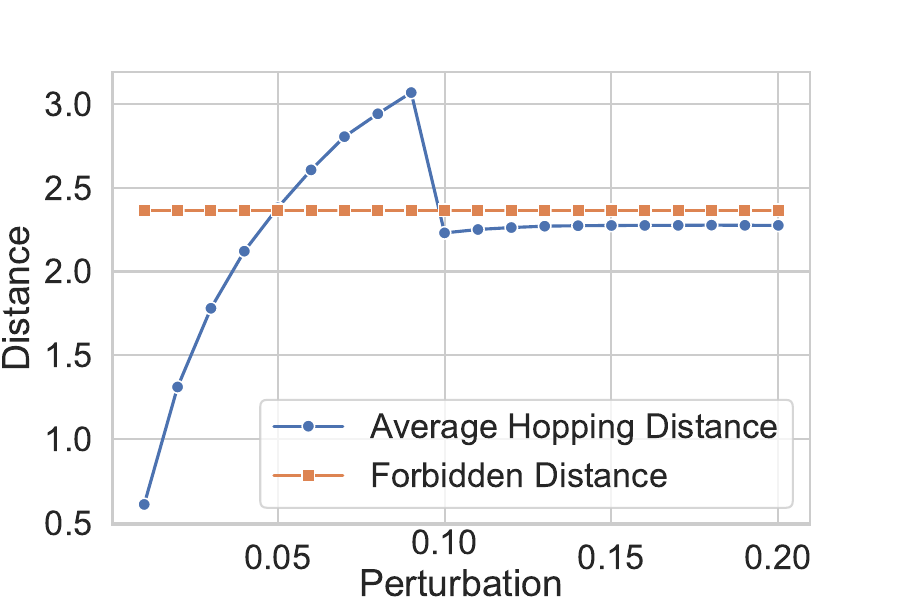}} 
	\subfloat[]{\includegraphics[trim=1 1 15 15,clip,width=0.25\linewidth,height=.23\linewidth]{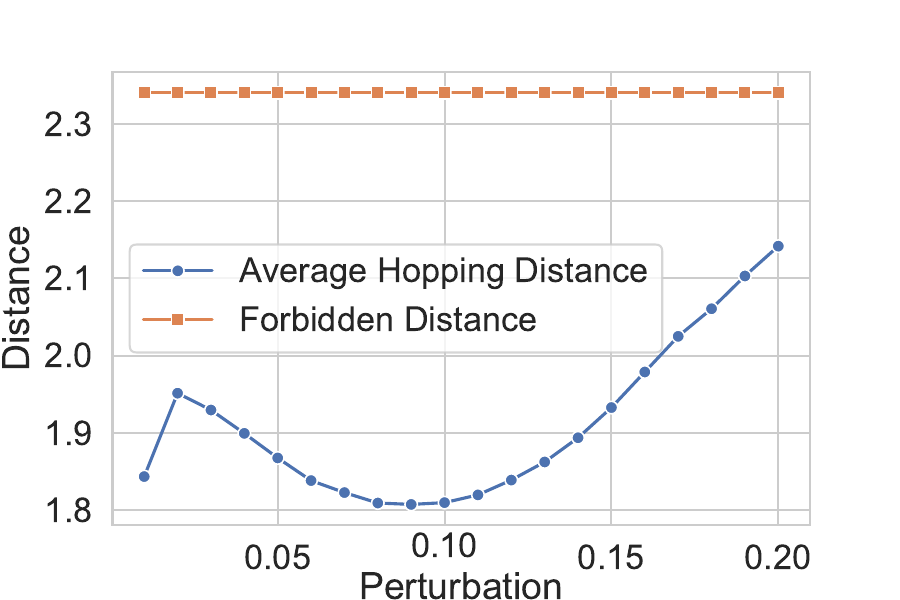}}
	\caption{The forbidden distance for a model and average hopping distance varying with perturbation $\epsilon$. (a) MNIST, (b) F-MNIST, (c) CIFAR10-ResNet32, and (d) CIFAR10-ImageNet} 
	\label{fig:EcUND} 
\end{figure} 
\begin{figure}
	\centering
	\subfloat[]{\includegraphics[trim=1 0 15 15,clip,width=0.25\linewidth, height= .25\linewidth]{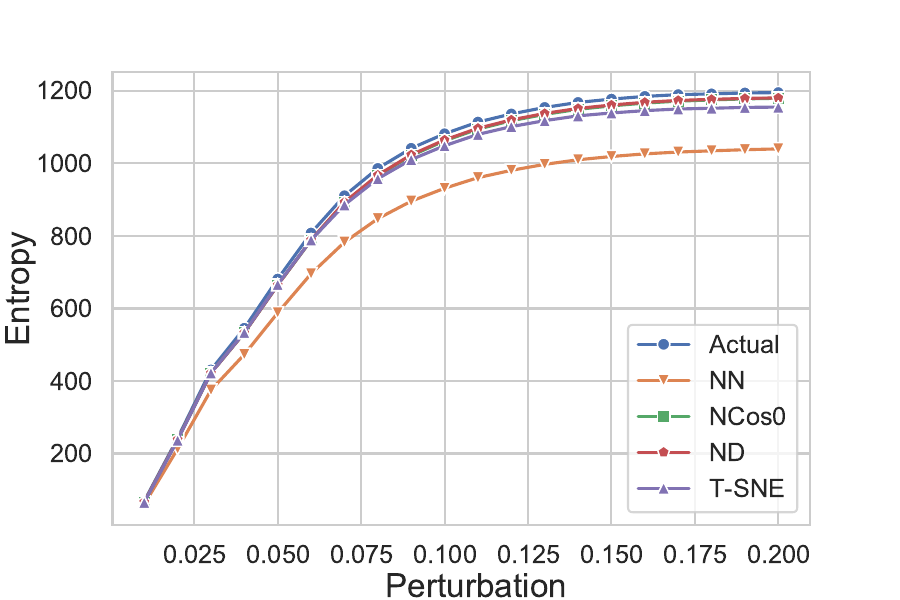}} 
	\subfloat[]{\includegraphics[trim=1 0 15 15,clip,width=0.25\linewidth, height=.25\linewidth]{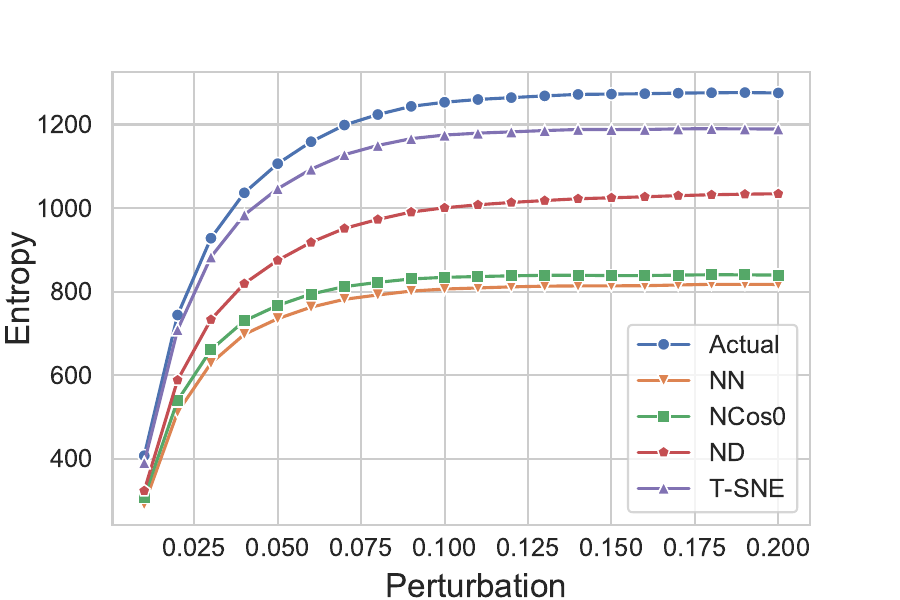}}
	\subfloat[]{\includegraphics[trim=1 0 15 15,clip,width=0.25\linewidth,height=.25\linewidth]{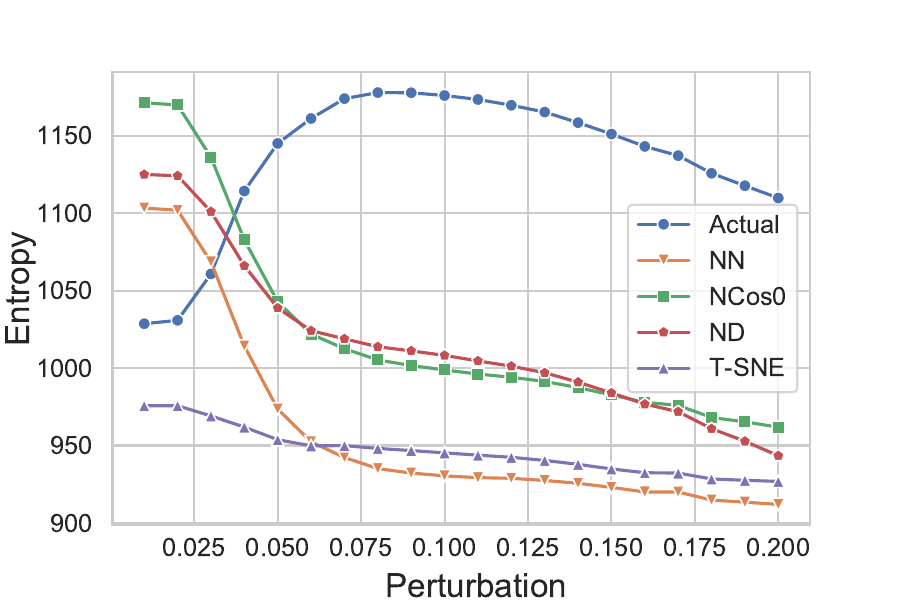}} 
	\subfloat[]{\includegraphics[trim=1 0 15 15,clip,width=0.25\linewidth, height=.25\linewidth]{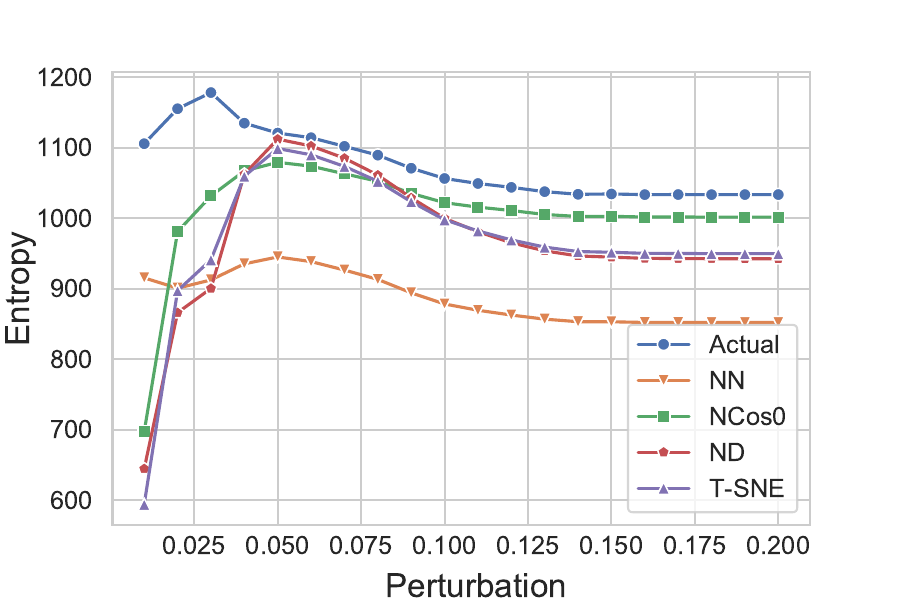}}
	\caption{Entropy varying with perturbation $\epsilon$ for a model prior applying our techniques and after applying each technique. (a) MNIST, (b) F-MNIST, (c) CIFAR10-ResNet32, and (d) CIFAR10-ImageNet} 
	\label{fig:entropy} 
\end{figure} 
\begin{figure}
	\centering
	\subfloat[]{\includegraphics[trim=5 1 15 15,clip,width=0.23\linewidth, height= .23\linewidth]{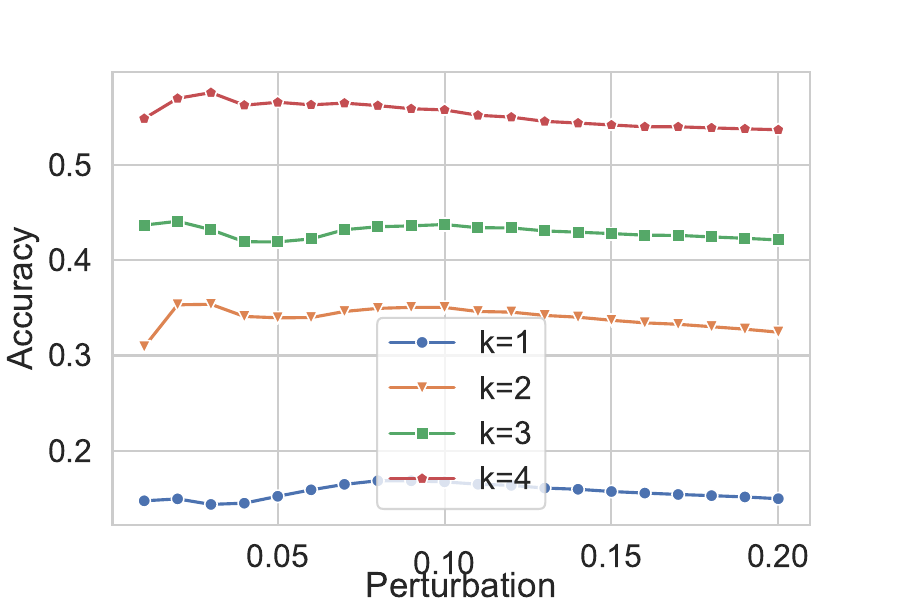}} 
	\subfloat[]{\includegraphics[trim=5 1 15 15,clip,width=0.25\linewidth, height=.23\linewidth]{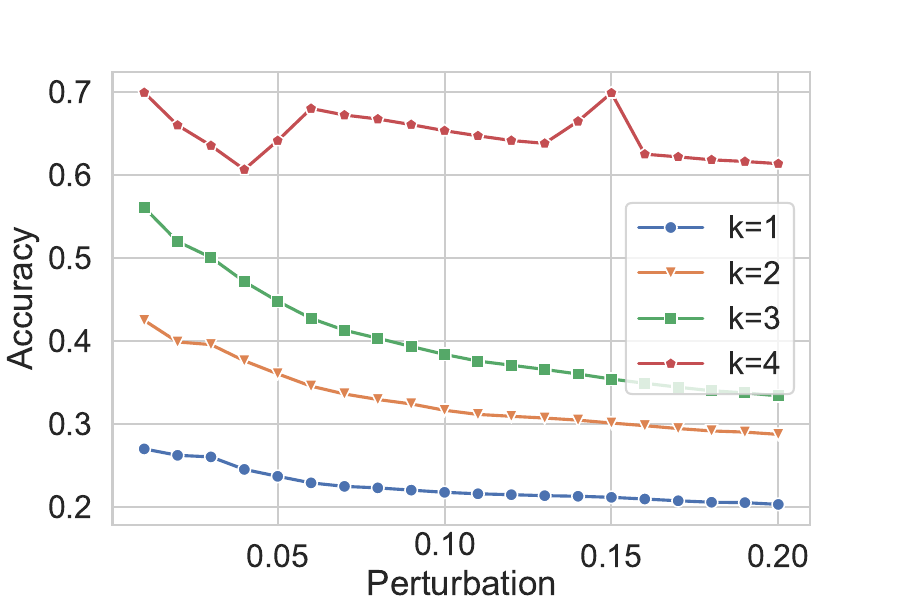}}
	\subfloat[]{\includegraphics[trim=5 1 15 15,clip,width=0.25\linewidth, height=.23\linewidth]{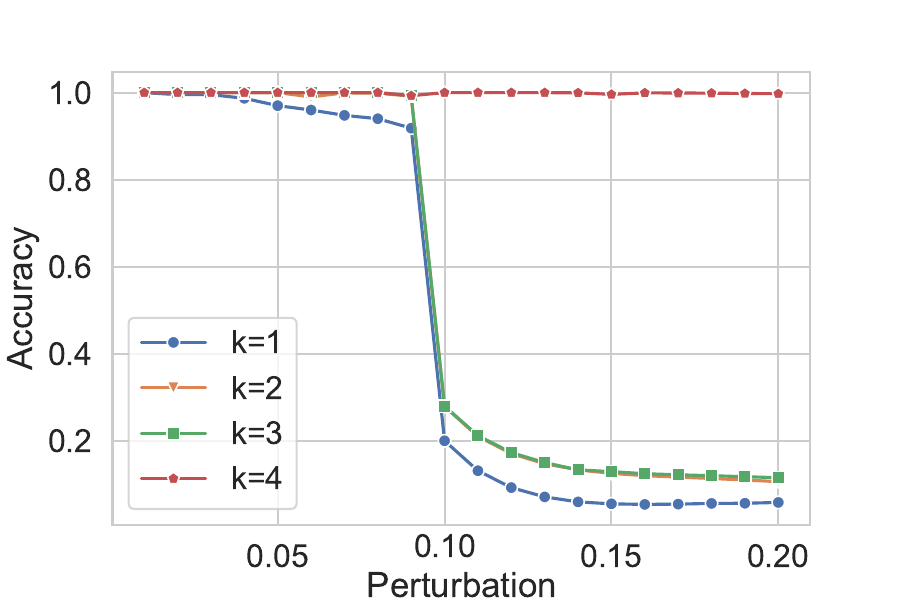}}
	\subfloat[]{\includegraphics[trim=5 1 15 15,clip,width=0.25\linewidth,height=.23\linewidth]{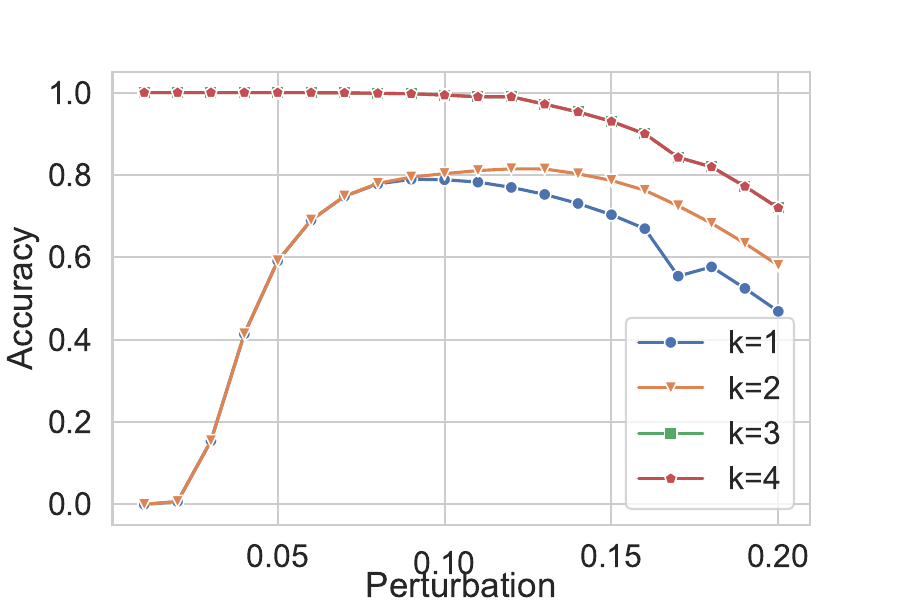}} 
	\caption{Accuracy varying with perturbation $\epsilon$ using (a) MNIST, (b) F-MNIST, (c) CIFAR10-ResNet32, and (d) CIFAR10-ImageNet .} 
	\label{fig:accuracy} 
	\vspace{-16pt}
\end{figure} 
\subsection{Usability of adversarial map for susceptible class detection}
\begin{wrapfigure}{r}{0.4\linewidth}
	\vspace{-15pt}
	\centering
	\subfloat[]{\includegraphics[width=0.3\linewidth]{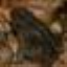}} 
	\subfloat[]{\includegraphics[width=0.3\linewidth]{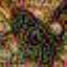}}
	\subfloat[]{\includegraphics[width=0.3\linewidth]{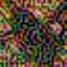}}
	\caption{(a) Actual image data from CIFAR-10, (b) Image with $\epsilon=0.1$, (c) Image with $\epsilon=0.2$.} 
	\label{fig:inverse} 
	\vspace{-10pt}
\end{wrapfigure}

We have claimed that using the \mapname we can identify the susceptible classes.  
We will discuss the accuracy of the best distance-based measure in \S\ref{subsec:accuracy}.
We have evaluated our approach on four separate models. 
In \S\ref{subsec:setup}, we have briefly described each model. 
For MNIST and F-MNIST, we have utilized a simple model with one input, 
one dense and one output layer.  
We have used state-of-the-art models for 
CIFAR-10 to evaluate our techniques. 
To calculate the distance, we have implemented four different measures 
and compared among them by computing the entropy of the model. 
Initially, we have calculated the entropy of a model, by applying an adversarial attack with a fixed perturbation. In the equation, $e_M=-\Sigma P(x_i)\log P(x_i)$, we assume that without any prior information, the probability of an input misclassified as an adversarial class given the actual class is $\frac{1}{n-1}$. For a fixed model, the value of $P(x_i)$ is constant for all data points based on the previous assumption. However, we have leveraged \mapname to provide weighted probability to each adversarial class based on the calculated distance between them. We have calculated the entropy based on the weighted probability using calculated distance from the actual class e.g., for actual class $x_i$  has neighbors $x_j$ and $x_k$. Here, $x_j$ is closer to $x_i$. In this scenario, $P(x_j|x_i)>P(x_k|x_i)$. Our goal is to reduce the entropy in a model with the distance-based measures. In Figure \ref{fig:entropy}, we evaluate the change of randomness by computing the entropy for all four models with four distance-based measures and compare them. In all the cases, \emph{Nearest Neighbor Hopping} distance based measure performs best in decreasing the entropy of a model under an adversarial setting. We have found that with increasing perturbation, the randomness typically increases. In contrast, the entropy in all the cases becomes more or less constant after a certain amount of perturbation. This indicates that, mostly all images are misclassified after certain perturbation and thus the entropy will not change in relation to the perturbation. Surprisingly, we have found that for CIFAR-10 ImageNet and ResNet-32 model, the entropy decreases with increasing perturbation.
In Figure \ref{fig:entropy}(c) and (d), initially the entropy increases with increasing perturbation but decreases with increasing perturbation after a certain simulation. We found that data points that have been misclassified with lower perturbation, were classified correctly with higher perturbation. In Figure \ref{fig:inverse}, initially the image has been classified correctly as \emph{frog} by the ImageNet model. With perturbation $\epsilon=0.1$, the image has been misclassified as \emph{deer}. Whereas, with perturbation $\epsilon=0.2$, the image has been classified correctly. 
\subsection{Effect of forbidden distance on adversarial attack}
In this section, we have shown that the forbidden distance for different attacks and models. Moreover, we have evaluated the effectiveness of $F_D$ in misclassification. We have claimed that the attack can not travel more than $F_D$ under a particular adversarial setting. We have proved our claim by computing $F_D$ on actual training data and demonstrate that the average hopping distance ($D$) traveled under an attack which remains less than $F_D$. After calculating $D$ using the Eq.\ref{eq:Hop}, we have simulated $D$ with increasing perturbation for each model. We have assumed that with increasing perturbation, the force of the attack increases so as the average distance (displacement) traveled by data points in $\mathbb{R}^N$. This is similar to the simple harmonic motion law of physics, which states that the displacement is proportional to the force. 
We have also evaluated the forbidden distance for each model and found whether the assumption regarding $D$ and $F_D$ applies. In Figure \ref{fig:EcUND}, we have simulated four cases and found that $D$ increases with perturbation and it remains under the bound given by $F_D$. Hence, $F_D$ provides an upper-bound distance for a model. But, in our approach, we have defined it as the capacity of withstanding an attack for a particular model.
\begin{table}[H]
	
	\vspace{-5pt}
	\centering
	\caption{Comparison of accuracy(\%) with learning based detection algorithms RCE\citep{3fawzi2018adversarial} and CE\citep{goodfellow2016deep} with NN(our approach).}
	\vspace{-5pt}
	\begin{tabular}{l|r|r|r|r|}
		\cline{2-5}
		& \multicolumn{4}{c|}{\textbf{Accuracy}}                                                                                                                                                                                                                             \\ \hline
		\multicolumn{1}{|l|}{\textbf{Method}} & \multicolumn{1}{l|}{\textbf{MNIST}} & \multicolumn{1}{c|}{\textbf{F-MNIST}} & \multicolumn{1}{c|}{\textbf{\begin{tabular}[c]{@{}c@{}}CIFAR-10 \\ ResNet\end{tabular}}} & \multicolumn{1}{c|}{\textbf{\begin{tabular}[c]{@{}c@{}}CIFAR-10\\ ImageNet\end{tabular}}} \\ \hline
		\multicolumn{1}{|l|}{\textbf{CE}}     & 79.7                                & -                                     & 71.5                                                                                     & -                                                                                         \\ \hline
		\multicolumn{1}{|l|}{\textbf{RCE}}    & 98.8                                & -                                     & 92.6                                                                                     & -                                                                                         \\ \hline
		\multicolumn{1}{|l|}{\textbf{NN}}     & 55.3                                & 64.8                                  & 99.9                                                                                     & 94.4                                                                                      \\ \hline
	\end{tabular}
	\label{tb:accuracy}
	\vspace{-5pt}
\end{table}
Using our approach, if the hopping distance between the classes $c_i$ and $c_j$ is more than $F_D$, then $c_i$ can not be misclassified to $c_j$. We have compared the model's $F_D$ with $D$ after a model has undergone through an attack. We have found that prior knowledge of a model provides a good estimation for describing the behavior of the adversarial examples.
\subsection{Effect of \mapname and susceptible class identifier}
\label{subsec:accuracy}
In this section, we take advantage of our \mapname and susceptible class identifier to analyze the threats to a DNN model. We have utilized the \mapname and computed the top $k$ susceptible target classes as described in the $\S\ref{subsec:lowcostdet}$. In Figure \ref{fig:accuracy}, we have simulated our approach varying perturbation and $k$. Though, it is apparent that with a larger value of $k$, the accuracy of predicting susceptible target classes will increase. We want to increase the accuracy of our approach with least value of $k$. This is a trade-off situation between $k$ and accuracy. From Figure \ref{fig:accuracy}, we have found that our approach performs best with $k=4$ for all models used in the evaluation. We have compared our work in Table \ref{tb:accuracy} with  reverse cross-entropy (RCE) \citep{3fawzi2018adversarial} and common cross-entropy (CE)\citep{goodfellow2016deep}. Our approach can identify the susceptible target class with higher accuracy using CIFAR-10 with ResNet-32 and ImageNet respectively. Whereas, the accuracy for DNN model using MNIST is lower than the previous work. To understand the reason, we have examined the adversarial classes for models using MNIST. We have found that model using MNIST has different \mapname for each actual class and all most all adversarial classes are susceptible to be attacked. To check further, we have visualized MNIST based model using t-SNE in 2-D space and have observed that the visualization shows a distinct separation among classes. Whereas, the 2-D visualization of the CIFAR-10 dataset based DNN model depicts some overlaps among the features, and our distance-based approach has discovered a certain pattern in the \mapname. Thus, we can conclude that our approach works better for models with high complexity e.g., CIFAR-10 based DNN models. 
\section{Conclusion}
\label{sec:conclusion}
In this paper, we have presented a technique to detect susceptible classes using 
the prior information of a model. 
First, we analyze a DNN model to compute the distance among classes in feature space. 
Then, we utilize that information to identify $k$ classes that are susceptible to 
be attacked. 
We found that with $k=4$, our approach performs best. 
To compare the four distance-based measures, we have presented a technique 
to create \mapname to identify susceptible classes. 
We have evaluated the utility of four different measures in creating \mapname. 
We have also introduced the idea of forbidden distance $F_d$ in the construction 
of \mapname. 
We have experimentally evaluated that the adversary can not misclassify to a target beyond distance $F_d$. 
We have found that \emph{Nearest Neighbor hopping} is able to describe the 
adversarial behavior by decreasing the entropy of a model and computing the 
upper bound distance ($F_D$) accurately. 
Our approach is also able to detect $k$ susceptible target classes that can 
detect adversarial examples with high accuracy for CIFAR-10 dataset 
(ImageNet and ResNet-32). 
In addition, for MNIST and F-MNIST, our approach possesses an accuracy 
of $55.3\%$ and $64.3\%$ respectively. 
Currently, our susceptible class detection identifies the source class of an 
adversarial example with probability. 
In the future, we want to find and study more properties of adversarial 
attack to detect adversarial examples with a lower bound guarantee. 
Analyzing model analysis techniques and algorithms to achieve the 
goal remain future work.


\bibliographystyle{plain}
\bibliography{neurips}

\end{document}